%% file: ogm_prediction.tex
\documentclass[10pt,twocolumn,letterpaper]{article}

\usepackage{cvpr}
\usepackage{times}
\usepackage{epsfig}
\usepackage{graphicx}
\usepackage{amsmath}
\usepackage{amssymb}
\usepackage{subfigure}
\usepackage{float}
\usepackage{multirow}
\usepackage[table,x11names]{xcolor}
\usepackage{array}
\usepackage{authblk}
\input{figures}

\usepackage[pagebackref=true,breaklinks=true,letterpaper=true,colorlinks,bookmarks=false]{hyperref}
\newcolumntype{x}[1]{>{\hspace{-2pt}\centering\arraybackslash}p{#1}}
 \cvprfinalcopy 

\def\cvprPaperID{4855} 

\ifcvprfinal\fi
\begin{document}

\title{Multi-Step Prediction of Occupancy Grid Maps with Recurrent Neural Networks}

\author{Nima Mohajerin}
\author{Mohsen Rohani\thanks{The authors contributed equally.}}
\affil{Huawei Noah's Ark, Markham, ON, Canada\\ \tt\small{\{nima.mohajerin, mohsen.rohani\}@huawei.com}\vspace{-10pt}}

\maketitle

\begin{abstract}
	We investigate the multi-step prediction of the drivable space, represented by Occupancy Grid Maps (OGMs), for  autonomous vehicles. Our motivation is that accurate multi-step prediction of the drivable space can efficiently improve path planning and navigation resulting in safe, comfortable and optimum paths in autonomous driving. We train a variety of Recurrent Neural Network (RNN) based architectures on the OGM sequences from the KITTI dataset. The results demonstrate significant improvement of the prediction accuracy using our proposed \textbf{difference learning} method, incorporating motion related features, over the state of the art. We remove the egomotion from the OGM sequences by transforming them into a common frame. Although in the transformed sequences the KITTI dataset is heavily biased toward static objects, by learning the difference between consecutive OGMs, our proposed method provides accurate prediction over both the static and moving objects. A video of the performance of our method on the KITTI dataset is available at \url{https://youtu.be/Bskd0Z7eLFE}.
\end{abstract}

\section{Introduction}
\label{sec:I}

\begin{figure}[t!]
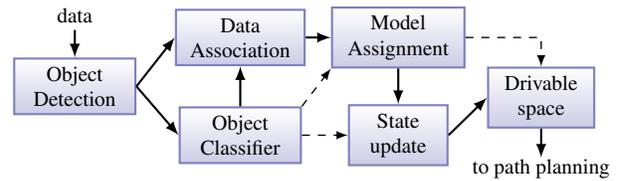
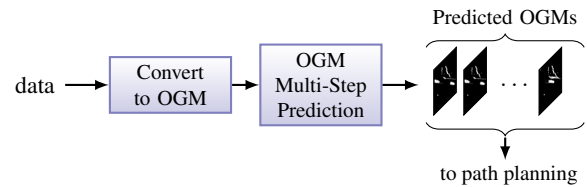

	\centering
	\subfigure[Classic approach for drivable space detection.]{\label{fig:classic}\ClassicOBJTracking\vspace{-10pt}}
	\subfigure[OGM multi-step prediction.]{\label{fig:OGM_MSP}\PredictionOBJTracking}
	\caption{Classic approach versus ours for providing the drivable space to a motion planner in autonomous driving.}
	\vspace{-5pt}
\end{figure}

Determining the environment state is a crucial ability for autonomous vehicles to have. Particularly for path planning and navigation, the state of the environment is required to determine safe areas to drive, that is, the drivable space. In a driving scenario, the classic approach is to detect and track objects, and based on the state of the tracked objects, determine the drivable space~\cite{Petrovskaya2009,Ess2010}. Figure~\ref{fig:classic} illustrates the general pipeline of the classic approach. As the new data comes in, the objects in the environment are detected (and classified). To keep track of objects, each object is given an ID. The Data Association module assigns existing IDs to the detected objects, or initiates new IDs if there is no matched tracks. The states of the tracked objects are updated using models which are assigned to them based on their class. Finally, the updated states (possibly along with the assigned models) are used to determine the drivable space.
 
The various stages in the classic approach require hand-engineered features.  Additionally, any error occurred at an early stage of the pipeline can propagate towards the next stages leading to an overall performance degradation. An alternative approach is to predict the drivable space at a level closer to the sensory information, requiring fewer abstraction levels. However, for such prediction to be useful, first a proper representation of the drivable space should be chosen which can be directly obtained from sensor measurements. One such proper means in this context is an Occupancy Grid Map (OGM)~\cite{Elfes1990, Tsardoulias2016}. An OGM divides the space around the ego-vehicle into equal cells which represent the occupancy state of surrounding regions, i.e., free or occupied. The state of the cells can be constructed using Laser range finders (and images~\cite{Santana2011}).  

Given an OGM, methods such as Probabilistic Road-Maps and Rapidly exploring Random Trees can be employed to generate a collision-free path~\cite{Tsardoulias2016}. In the presence of dynamic objects, however, a single OGM is insufficient for guaranteeing a safe passage in general, as it does not properly represent the moving objects. Planning based on the updated OGMs results in inefficient and uncomfortable paths requiring frequent steering corrections or even inadmissible paths. In such scenarios, one common approach is to ``grow'' the occupied regions by padding the occupied cells, which artificially creates a safe distance from obstacles. However, the amount of padding is difficult to determine a priori and depends on the objects' type, velocity, etc. Without proper object detection and classification, the padding size has to be chosen above a safe threshold which may result in severely restricting the ego-vehicle movement in crowded areas, such as urban environments, and failing to detect useful drivable spaces in highway scenarios where the objects may move relatively fast.

Alternatively, it is possible to provide multi-steps of OGM prediction into the future~\cite{Dequaire2018, Engel2018, Hoermann2017}. Multi-step prediction of OGMs in a dynamic environment provides the drivable space for a planning algorithm without the need for the several stages required in the classic approach. The idea is to encode the observed states of the environment, represented as OGMs, into latent variables of a recursive model, from which the future OGMs can be produced recursively. Recurrent Neural Networks (RNNs) are, therefore, a potential model candidate, which have also been successfully applied to various other problems with dynamic nature, such as unsupervised video prediction~\cite{Lotter2016, Finn2016}, multi-step prediction of mobile robots~\cite{Mohajerin2017_2} and visual odometry~\cite{Wang2018}. Moreover, it is shown that RNNs can produce promising results in the context of OGM prediction~\cite{Dequaire2018, Engel2018}.

OGMs may also be looked upon as black-and-white images. However, predicting images in a video stream, i.e., video prediction~\cite{Finn2016, Mathieu2015, Srivastava2015}, can be studied either as a regression problem, or multinomial classification (with at least 256 classes for gray-scale images), where OGM prediction is a binary classification problem. As a sanity check, we employed PredNet~\cite{Lotter2016} and the CDNA model ~\cite{Finn2016}, but both methods failed to provide a reliable OGM prediction. Particularly, the occlusions that occur within OGMs, due to the nature of range finder sensors, are difficult to be dealt with in the video prediction task where the primary goal is to predict the future images as similar as possible to the ground truth. Therefore, in this work we do not consider the multi-step prediction of OGMs as a video prediction task.

In \cite{Dequaire2018}, a network of Gated Recurrent Units~\cite{Cho2014} (GRUs), with dilated convolutions, is used whose states are initialized as it receives and reconstructs observed OGMs, for a limited number of steps. Then, the initialized states are employed to generate next OGMs, while the input OGMs are blank. Inspired by \cite{Dequaire2018}, we propose an approach to learn the \emph{difference} between consecutive OGMs as a compensation matrix. The current OGM, modified by the compensation matrix, provides the features for a predictive classification. This choice stems from the fact that the current OGM provides a reliable prior for the drivable space. The major source of discrepancy between consecutive OGMs is the egomotion, which presumably is available (e.g., from the Drive-By-Wire system or GPS). Therefore, in this work, the observed OGMs are transferred to a common frame using egomotion information. The common frame corresponds to the ego-vehicle coordinate system at the present time, i.e., the frame at which we want to predict future OGMs.

The OGMs are also predicted into the future with respect to the common frame. In fact, in our approach the drivable space is predicted as OGMs in a frame which is absolute over the observed history and the prediction period, but is attached to the ego-vehicle. This is different to the approach taken in~\cite{Dequaire2018} where a Spatial Transformer Module (STM) ~\cite{Jaderberg2015} is employed to take account for the egomotion. Since our intention is to predict OGMs suitable for planning, it is reasonable to predict the OGMs regardless of the \emph{future} egomotion. Providing the environment state in an absolute frame makes it possible for a planning algorithm to examine and plan based on virtually all ego trajectories that are safe, i.e., those which do not interfere with other objects' paths in the environment.

Training a network for OGM prediction is supervised as the target output is the OGM. However, the targets directly come from the sensors and do not need human supervision to be generated. Such datasets are easy to collect which significantly reduces the human labor and improves the time-to-market. One difficulty is that the number of static objects is quite often dominant in OGM datasets. Consequently, the network can easily trap in a local minimum where it assumes the surrounding environment is basically static. Learning the difference between OGMs partially prevents such local minima as the network only needs to compensate for the moving objects. Additionally, we propose to incorporate motion relevant information as inputs to the networks, either by using standard optical-flow extraction algorithms such as the Farneback method~\cite{Farneback2003}, or by feeding the OGM differences. Both methods are considered and evaluated for performance and computation time.

Our contributions are summarized as follows,
\begin{itemize}
	\item A novel RNN-based architecture is proposed to predict OGMs over multi-steps into the future, based on learning the OGM differences between consecutive frames,
	\item Motion related features are employed which enhances OGM prediction over dynamic objects significantly,
	\item A conclusive comparison between various RNN-based architectures is presented for the problem at hand.
\end{itemize}

In the following, relevant works are reviewed in Section~\ref{sec:II}. Section~\ref{sec:III} formulates the OGM prediction problem and describes our methods to address it. Section~\ref{sec:IV} presents the experimental results and discussions. We conclude the paper in Section~\ref{sec:V}.

\section{Related Work}
\label{sec:II}
In \cite{Ondruska2016}, the authors propose a deep tracking scheme, where a simple RNN is leveraged to learn OGM-to-OGM mappings. The training data consists of OGMs that are built by synthesized laser readings collected from a static sensor. This work is improved and expanded in \cite{Dequaire2018}, where the egomotion is handled with STMs, and convolutional GRUs, with dilation, replace the vanilla RNN architecture. However, the two approaches in \cite{Ondruska2016} and \cite{Dequaire2018} are similar: a sequence of observed OGMs, followed by blank OGMs, are fed to an RNN, whose output is trained on the observed OGMs for the entire sequence. Therefore, the RNN acts as an autoencoder while it receives the observed OGM, to initialize its states. The accumulated information in the states is employed to predict the OGM evolution as the network receives blank OGMs. However, feeding blank inputs cause the output OGMs to fade out rapidly. Also, approaches in \cite{Ondruska2016} and \cite{Dequaire2018} become computationally inhibitive as the OGM size increases unless the OGMs are either resized or an encoder-decoder architectures is employed.

Authors in \cite{Hoermann2017} and \cite{Engel2018} employ Dynamic OGMs (DOGMa). DOGMa is the result of fusing a variety of sensor readings using Bayesian filtering, which associates dynamic information to each cell as well as the occupancy state~\cite{Nuss2018}. The dynamic information contains the velocity and its uncertainty. An encoder-decoder structure with Convolutional Long-Short-Term-Memory network (ConvLSTM)~\cite{Xingjian2015} receives DOGMa and produces the occupancy probability of static regions alongside the anchor boxes for dynamic objects \cite{Engel2018}. An automatic output label generation is also used which can have a ``relatively high false negative rate"~\cite{Hoermann2017}. Additionally, the datasets in \cite{Hoermann2017} and \cite{Engel2018} are collected from a static sensor (parked car).

OGM prediction has already been used in path planning of autonomous robots. In~\cite{Noguchi2012}, Variable Length Markov Model is employed to predict the OGM in an environment whose main occupants are humans. However, in ~\cite{Noguchi2012}, the experiments are restricted to cases where there is only one human exists in the vicinity of the autonomous robot. Similarly in \cite{Ohki2010}, multi-steps of predicted OGMs are stacked in an XYT space (T stands for time) which is then used for path planning of rescue robots. The XYT stack of predicted OGMs is also used in other works, such as \cite{Gupta2008}, however, the OGMs are either updated using object models (classic approach) or the main occupant are of the same object type, e.g. human in~\cite{Noguchi2012} and \cite{Ohki2010}.

\section{OGM Prediction in Autonomous Driving}
\label{sec:III}

In practice, an OGM cell can be either fully occupied, partially occupied or completely free. However, in this work, it is assumed that a cell can be either fully occupied or free, with a probability of occupancy associated to it. Formally, the state of the cell located at $i$\textsuperscript{th} row and $j$\textsuperscript{th} column of the OGM, at time-step $k$, is represented by a binary random variable $c_k(i,j)\in\{0,1\}$. For now, assume that the OGM at time-step $k$, $\mathbf{O}_k$, is the probability matrix of $c_k(i,j)$, that is,
\begin{equation}
	\mathbf{O}_k=\Big{[}p\big{(}c_k(i,j)\big{)} \Big{]}=
		\begin{bmatrix}
			p\big{(}c_k(1,1)\big{)} & \dots & p\big{(}c_k(1,X)\big{)} \\
			\vdots & \ddots & \vdots \\
			p\big{(}c_k(Y,1)\big{)} & \dots & p\big{(}c_k(Y,X)\big{)} \\
		\end{bmatrix},
	\label{eq:ogm0}
\end{equation}
where $X$ and $Y$ indicate the grid size. In order not to lose generality, we assume that at each time-step, $c_k(i,j)$s, where $i=1,...,Y$ and $j=1,...,X$, are independent with possibly different distributions. However, $c_k(i,j)$s are dependent over time. In general, the probability of $c_k(i,j)$ is conditioned on the set of random variables $C_{i,j}$,
\begin{equation}
	\begin{aligned}
		C_{i,j}=\big{\lbrace}c_v(m,n)|&v=k-1,k-2,...\\
								&m=i-\alpha_i,...,i+\beta_i,\\
								&n=j-\alpha_j,...,,j+\beta_j,\\
						   		&0\leq\alpha_i < i, 0\leq\beta_i\leq X-i,\\
						   		&0\leq\alpha_j\leq j,0\leq\beta_j<Y-j\big{\rbrace},
	\end{aligned}
	\label{eq:C}
\end{equation} 
where $\alpha_i, \beta_i$ and $\alpha_j, \beta_j$ are integers and indicate some neighborhood around $i$ and $j$. In general, the values $\alpha_i, \beta_i$ and $\alpha_j, \beta_j$ depend on a number of factors, such as the ego-vehicle's velocity and/or the speed of the surrounding objects. Therefore, we assign the extreme possible values to them, in which case, $(m,n)$ covers the entire OGM. Therefore, the elements of OGM matrix in \eqref{eq:ogm0} are, in fact, conditional probabilities,
\begin{equation}
	p\big{(} c_k(i,j)|\mathcal{C}_{k-1},\mathcal{C}_{k-2},...\big{)},
	\label{eq:probs}
\end{equation}
where,
\begin{equation}
	\mathcal{C}_k=\big{\lbrace}c_k(m,n)|m=1,...,Y;n=1,...,X\big{\rbrace},
\end{equation}
and we define the OGM as the matrix of the \emph{conditional} probabilities~\eqref{eq:probs},
\begin{equation}
	\begin{aligned}
		\mathbf{O}_k=&\Big{[}p\big{(} c_k(i,j)|\mathcal{C}_{k-1},\mathcal{C}_{k-2},...\big{)} \Big{]},\\
		&i=1,...,Y; j=1,...X.
	\end{aligned}
	\label{eq:ogm}
\end{equation}

Our goal is to predict the OGMs, as depicted in \eqref{eq:ogm}, over multi-steps into the future. To this end, we employ and train a variety of RNN-based architectures. The feedback connections in an RNN are particularly useful to establish the recursive dependency represented in \eqref{eq:ogm}. The RNN-based  models in this work are employed as sequence-to-sequence maps, i.e., they receive a sequence of input OGMs and produce a sequence of OGMs, of the same length. Therefore, given a fixed sequence length, $T$, a \emph{prediction task} is defined as feeding the input OGM sequence, $\mathbb{O}$, to an RNN-based model which generates the output OGM sequence, $\mathbb{\hat{O}}$,
\begin{subequations}
	\begin{align}
		\mathbb{O}&=\{\mathbf{O}_k\}, k=1,...,T,
		\label{eq:inputOGM}\\
		\mathbb{\hat{O}}&=\{\mathbf{\hat{O}}_k\}, k=1,...,T,
		\label{eq:outputOGM}
	\end{align} 
	\label{eq:seq2seq}
\end{subequations}
where $\mathbf{\hat{O}}_k$ is the model output at time-step $k$ whose form depends on the model architecture. Based on~\cite{Dequaire2018} we initially form the input sequence as follows,
\begin{equation}
	\mathbf{O}_k=\begin{cases}
		\mathbf{O}^*_k, k=1,...,\tau_I,\\
		[\mathbf{0}], k=\tau_I+1,...T,
	\end{cases}
	\label{eq:OCases}
\end{equation}
where $\mathbf{O}^*_k$ is the \emph{observed} OGM at $k$ and $[\mathbf{0}]$ represents a matrix of zeros whose size is the same as $\mathbf{O}^*_k$, i.e., a blank OGM. Later we will modify the input sequence. Equation~\eqref{eq:seq2seq} and \eqref{eq:OCases} indicate that a prediction task consists of two phases. During the first phase, where $k=1,...,\tau_I$, the observed OGMs are given to the model. The first phase is mainly intended for RNN state initialization~\cite{Mohajerin2017_1}, and therefore, we will refer to it as the initialization phase, or init-phase for short. The second phase, namely the prediction phase, starts at $k=\tau+1$ and is intended for multi-step prediction, as the input OGMs are blank. The length of the prediction phase is $\tau_P=T-\tau_I$.

It is worthwhile to mention that in~\cite{Dequaire2018}, authors train their network to reconstruct the input during the init-phase, therefore, their architecture behaves similar to an autoencoder. From our point of view, such approach does not necessarily result in extracting features that are useful for prediction, rather, those features are useful for reconstructing the inputs. In contrary, our models are trained to generate one-step-ahead prediction during the init-phase. Therefore, we force the models to learn features that are useful for prediction.

\subsection{Base Architecture}
\label{sec:basearch}
\begin{figure}[h]
	\centering
	\includegraphics[scale=0.55,trim=22mm 8mm 25mm 10mm,clip]{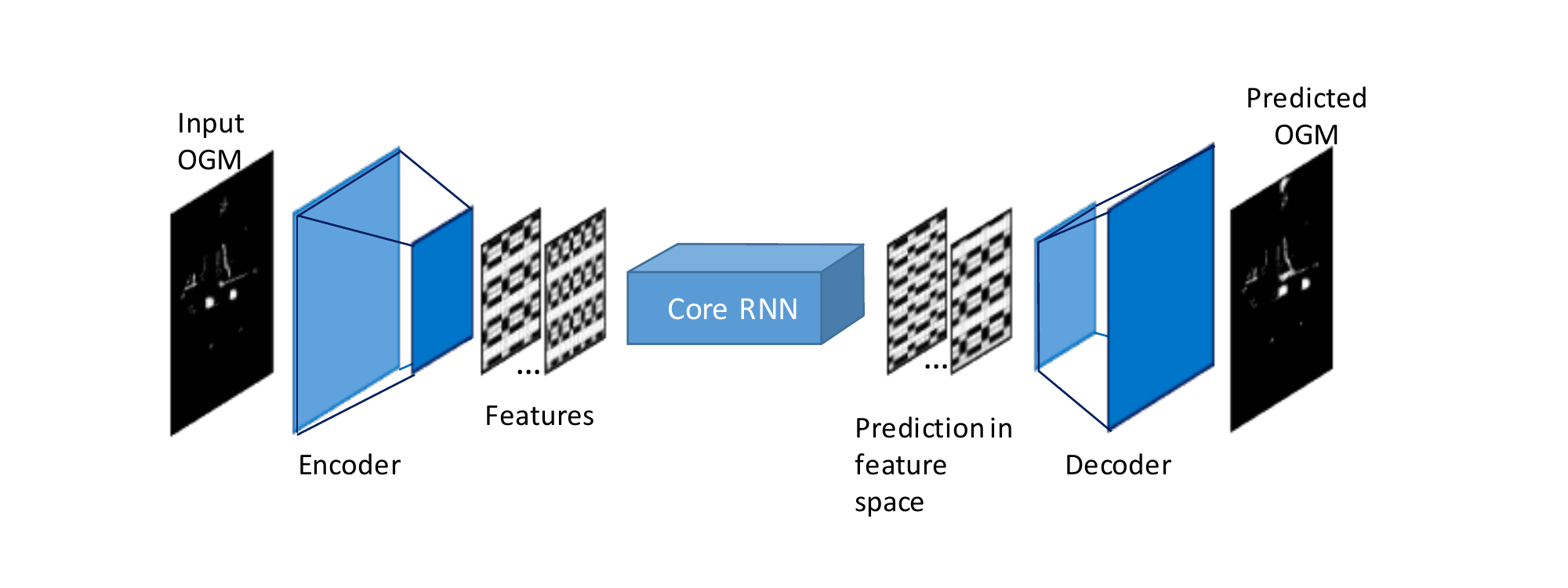}
	\caption{Base Architecture.}
	\label{fig:BaseArch}
	\vspace{-10pt}
\end{figure}

To handle large OGMs, an encoder/decoder architecture is employed, as depicted in Figure~\ref{fig:BaseArch}. The encoder module consists of a number of convolutional layers, each followed by a max pooling, to reduce the input OGM size and extract features. Then, the features are passed through an RNN, followed by a decoder module. The decoder module upsamples the RNN output using transposed convolutional layers. The number of layers in the encoder and the decoder is the same. Also, the decoder's output has the same XY dimension as the input OGM.

To account for the egomotion, the OGMs are transferred to the frame at which the prediction starts, i.e., $k=\tau$. We use standard geometric image transformation using the displacement in $x$, $y$ and the heading directions over consecutive frames. This transformation is applied to all of the OGMs in each prediction task. As a result, the OGMs (observed and predicted per each prediction task) are treated as if they are seen from a stationary sensor whose location is the same as the sensor location at $k=\tau$. To avoid notation clutter, we use the same notation for the OGMs as before, keeping in mind that they are all transformed accordingly.

\subsection{Extended Architectures}
We extend the base architecture by feeding the output back to the network. Therefore, \eqref{eq:OCases} becomes,
\begin{equation}
\mathbf{O}_k=\big{[}1-u(k-\tau-1)\big{]}\mathbf{O}^*_k+u(k-\tau-1)\mathbf{\hat{O}}_k, 
\label{eq:OCases_of}
\end{equation}
where $\mathbf{\hat{O}}_k$, is the predicted OGM at time-step $k$ and $u(.)$ is the standard step function\footnote{$u(x)$ returns 1 for $x\geq0$ and 0 otherwise.}. The output feedback connection incorporates the information, that are coded inside the decoder module, back into the network.

\begin{figure}[h]
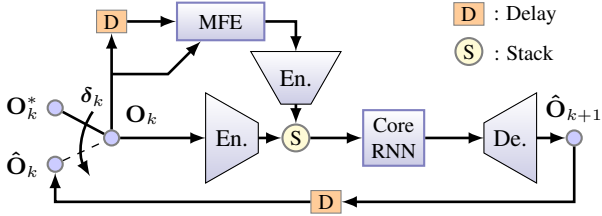

	\centering
	\BaseArchExtensions
	\caption{Extensions over the base architecture. ``MFE" stands for Motion-related Feature Extraction. The switch $\boldsymbol{\delta}_k$ changes position at $k=\tau_I+1$. ``En." and ``De." stand for Encoder and Decoder, respectively.}
	\label{fig:MFRNN}
\end{figure}

The second extension is to include motion-related features using Motion-Flow (MF) extraction algorithms (e.g., Farneback algorithm~\cite{Farneback2003}), or a \emph{two-channel difference} method. An MF extraction algorithm, receives two consecutive OGMs and provides a tensor of the same size and depth of two. Each of the two channels correspond to the movement in X and Y directions. The two-channel difference method takes the difference between two past consecutive OGMs and places the result into two channels; one channel for the positive values and the other for (the absolute value of) the negative values. The resulting two-channel tensor, from either of the methods, is then passed through an encoder module, whose output is stacked with the output of the OGM encoder. The extended architectures is illustrated in Figures~\ref{fig:MFRNN}.

\subsection{Difference Learning Architecture}
With the current technology, high resolution OGMs can be constructed at 10 Hz rate and faster. In a normal driving scenario, it is quite unlikely that the consecutive OGMs at such rates differ dramatically. In other words, current OGM provides a reasonable prior for predicting the next OGM. 

\begin{figure}[h]
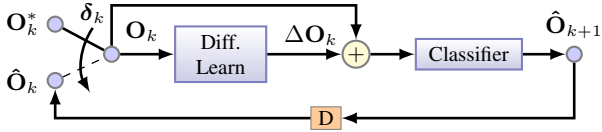

	\centering
	\DiffLearningBlockDiagram
	\caption{Difference OGM learning model.}
	\vspace{-10pt} 	
	\label{fig:DLModel}
\end{figure}

\begin{figure}[ht!]
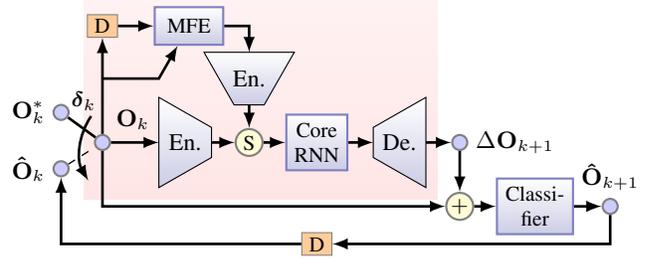

	\centering
	\DiffLearningBlockDiagramFull
	\caption{Detailed architecture for the difference learning model. The shaded area is the \textbf{Diff. Learn} module.}
	\label{fig:DLModelFull}
	\vspace{-10pt}
\end{figure}

We embed the above idea into a \emph{difference learning} architecture, which is based on a two-stage process as illustrated in Figure~\ref{fig:DLModel}. The first stage, depicted by \textbf{Diff. Learn}, generates a \emph{compensation matrix}, $\Delta\mathbf{O}_k$, whose size is the same as the input OGM. The \textbf{Diff. Learn} module employs the extended architecture depicted in Figure~\ref{fig:MFRNN}. The result of adding $\Delta\mathbf{O}_k$ to the input OGM provides the basis on which the \textbf{Classifier} module produces the next OGM prediction. Therefore, the \textbf{Diff. Learn} module should implicitly distinguish between the static and dynamic objects which is then reflected in $\Delta\mathbf{O}_k$. The elements of $\Delta\mathbf{O}_k$ are real values in $[-1, 1]$. A value near zero corresponds to a cell whose occupancy should not be altered, i.e., a free cell or a cell occupied by a static object). Similarly, a value closer to 1 (or -1) attempts to add (or clear) occupancy to (from) the corresponding cells, indicating the cell is being occupied by (or freed from) a dynamic object. The \textbf{Classifier} module refines the modified OGM further to predict the next OGM, i.e., $\mathbf{\hat{O}}_{k+1}$. Figure \ref{fig:DLModelFull} illustrated the detailed difference-learning model. Note that the \textbf{Classifier} module can be a simple feed-forward or a recurrent network. 

\subsection{Training Cost}
As the problem at hand is classification (occlusion vs. free), we employ pixel-wise Cross-Entropy cost between the output, $\mathbf{\hat{O}}_{k}$, and the target, $\mathbf{O}^*_{k}$. Because the majority of the cells are normally unoccupied, the cost is normalized with respect to the number of occupied/free cells. Also, the pixel-wise cost is multiplied by the visibility matrix as in \cite{Dequaire2018}, to let the network handle occlusions.

To make sure the \textbf{Diff. Learn} module is effectively employed, two extra terms are added for the difference learning models. First, based on the idea that most of the objects are static in a driving scenario, an $L_2$ norm on the compensation matrix, $\Delta\mathbf{O}_{k+1}$, is added to the total cost. This term ensures that the non-zero elements inside $\Delta\mathbf{O}_{k+1}$ correspond to the cells whose occupancy should be changed during the prediction task, i.e., the dynamic objects. To force the classifier inputs towards the target, a multi-scale Structural Similarity Index Metric \cite{Wang2003} (SSIM) between the summation result and the target OGM is added to the cost.

\begin{table*}
	\footnotesize
	\begin{center}
		\begin{tabular}{|c|c|c||c|c|c|c|c|c|}
			\hline
			\# & Model & Cfg. & Arch. & Input & MFE  & Di. & \textbf{Diff. Learn} & \textbf{Classifier}\\
			\hline
			1 & \multirow{3}{*}{ED} & Base & Fig.~\ref{fig:BaseArch}& Eq.~\eqref{eq:OCases}& N/A &  No & N/A & N/A\\
			\cline{1-1}	\cline{3-9}
			2 & & Ext1. & Fig.~\ref{fig:MFRNN} & Eq.~\eqref{eq:OCases_of}& Farenback &  No & N/A & N/A\\
			\cline{1-1}	\cline{3-9}
			3 & & Ext2. & Fig.~\ref{fig:MFRNN} &  Eq.~\eqref{eq:OCases_of}& Two-channel difference &  No & N/A &N/A\\
			\hline
			4 & \multirow{3}{*}{ED-Di} & Base& Fig.~\ref{fig:BaseArch}& Eq.~\eqref{eq:OCases}& N/A &  Yes & N/A & N/A\\
			\cline{1-1}	\cline{3-9}
			5 & & Ext1. & Fig.~\ref{fig:MFRNN} & Eq.~\eqref{eq:OCases_of}& Farenback &  Yes & N/A & N/A\\
			\cline{1-1}	\cline{3-9}
			6 & & Ext2. & Fig.~\ref{fig:MFRNN} &  Eq.~\eqref{eq:OCases_of}& Two-channel difference &  Yes & N/A &N/A\\
			\hline
			7 & \multirow{3}{*}{Diff.1} & Base & Fig.~\ref{fig:DLModel}& Eq.~\eqref{eq:OCases_of} & N/A &  No & Base (Fig.~\ref{fig:BaseArch}) & 2 Layer Conv. \\
			\cline{1-1}	\cline{3-9}
			8 & & Ext1. & Fig.~\ref{fig:DLModelFull}& Eq.~\eqref{eq:OCases_of} & Farenback  &  No & Shaded area in  Fig.~\ref{fig:DLModelFull} & 2 Layer Conv. \\
			\cline{1-1}	\cline{3-9}
			9 & & Ext2.& Fig.~\ref{fig:DLModelFull}& Eq.~\eqref{eq:OCases_of} &Two-channel difference  &  No & Shaded area in  Fig.~\ref{fig:DLModelFull} & 2 Layer Conv. \\
			\hline
			10 & \multirow{3}{*}{Diff.2} & Base & Fig.~\ref{fig:DLModel}& Eq.~\eqref{eq:OCases_of} & N/A &  No & Base (Fig.~\ref{fig:BaseArch}) & 1 ConvLSTM \\
			\cline{1-1}	\cline{3-9}
			11 & & Ext1. & Fig.~\ref{fig:DLModelFull}& Eq.~\eqref{eq:OCases_of} & Farenback  &  No & Shaded area in  Fig.~\ref{fig:DLModelFull} & 1 ConvLSTM \\
			\cline{1-1}	\cline{3-9}
			12 & & Ext2.& Fig.~\ref{fig:DLModelFull}& Eq.~\eqref{eq:OCases_of} &Two-channel difference  &  No & Shaded area in  Fig.~\ref{fig:DLModelFull} & 1 ConvLSTM \\
			\hline
		\end{tabular}
	\end{center}
	\vspace{-5pt}
	\caption{The models trained in this work. ED models constitute the base line over each configuration (Cfg.), ED-Di models are the closest replica of the model proposed in ~\cite{Dequaire2018}, in our framework, Diff.1 and Diff.2 models represent our proposed models.}
	\label{tab:Models}
\end{table*}
\normalsize

\section{Experimental Results}
\label{sec:IV}

The OGM sequences are obtained from the KITTI raw dataset~\cite{Geiger2013}. The raw Lidar point-clouds, collected at 10Hz, are converted to bird's eye view OGMs (with ground removal) using the method described in~\cite{Miller2007}. Then, the OGM sequences are partitioned into segments of 20 frames, preserving the order within each segment. Every segment corresponds to 2 seconds of OGM measurement and is regarded as one sample. The init-phase length is set to 10 steps, i.e., $\tau=10$. Since the models' ability to handle occlusion is important to us, in addition to the natural occlusions that exist in the raw data, we impose a constant occlusion pattern over the OGMs by filtering out occupancies outside the front camera field of view (FOV). With a cell size equal to $20$(cm)$\times20$(cm), each OGM frame consists of $256\times256$ cells, arranged in a square matrix, corresponding to occupancies over 50 meters in front of the car. The OGMs are visualized as $256\times256$ pixel, black and white or gray-scale, images, where the former shows the actual occupancy and the latter illustrates the discretized occupancy probabilities. Each pixel represents one OGM cell.

Table~\ref{tab:Models} lists the models, and their configurations, trained for this work. The ED models provide a basis for evaluation. The ED-Di models correspond to the model proposed in~\cite{Dequaire2018} along with two improvements based on using the motion-related features. The Diff.1 and Diff.2 models represent our proposed difference learning architecture.

\begin{table*}
	\footnotesize
	\begin{center}
		\begin{tabular}{|c|x{0.6cm}||ccc|ccc||ccc|ccc||c||c|}
			\hline
			\multicolumn{2}{|c||}{Architectures} & \multicolumn{6}{c||}{Whole OGM} & \multicolumn{6}{c||}{Objects only} & \multirow{3}{0.4cm}{t/f (ms)} & \multirow{3}{0.4cm}{Ord.} \\
			\cline{1-14}
			\multirow{2}{*}{Model} & \multirow{2}{*}{Cfg.} & \multicolumn{3}{c|}{Whole seq.} & \multicolumn{3}{c||}{Prediction seq.} & \multicolumn{3}{c|}{Whole seq.} & \multicolumn{3}{c||}{Prediction seq.} & & \\
			\cline{3-14}
			& & TP & TN & S\textsubscript{100} & TP & TN & S\textsubscript{100} & TP & TN & S\textsubscript{100} & TP & TN & S\textsubscript{100} & &\\
			\hline
			& Base & 87.51 & 97.85 & 93.91 & 85.37 & 97.38 & 92.95 & 77.07 & 97.88 & 95.71 & 70.39 & 97.56 & 95.01 & \textbf{3.8} & 11 \\
			& Ext.1 & 87.87 & 87.29 & 93.75 & 85.25 & 99.07 & 95.49 & 77.19 & 98.94 & 96.79 & 70.28 & 98.59 & 96.02 & 9.2 & 12 \\
			\rowcolor[gray]{0.8}\multirow{-3}{*}{\cellcolor{white}ED}& \cellcolor{white}Ext2. & 89.48 & 98.69 & 94.85 & 87.50 & 98.21 & 94.31 & 78.95 & 98.77 & 96.55 & 72.76 & 98.39 & 95.77 & \cellcolor{white}4.2 & \cellcolor{white}5\\
			\hline
			& Base & 86.16 & 97.41 & 93.89 & 86.01 & 97.68 & 93.10 & 77.83 & 98.01 & 95.91 & 71.84 & 97.89 & 95.14 & 4.0& 9 \\
			& Ext.1 & 88.09 & 89.81 & 94.02 & \cellcolor[gray]{0.8}85.89 & \cellcolor[gray]{0.8}99.10 & \cellcolor[gray]{0.8}96.92 & \cellcolor[gray]{0.8}78.24 & \cellcolor[gray]{0.8}98.55 & \cellcolor[gray]{0.8}97.01 & \cellcolor[gray]{0.8}71.25 & \cellcolor[gray]{0.8}98.88 & \cellcolor[gray]{0.8}\textbf{97.03} & 9.8 & 7 \\
			\multirow{-3}{*}{ED-Di}& Ext.2 & \cellcolor[gray]{0.8}88.60 & \cellcolor[gray]{0.8}97.90 & \cellcolor[gray]{0.8}94.57 & 87.15 & 97.52 & 95.52 & 76.52 & 96.26 & 95.25 & 69.42 & 97.93 & 93.87 & 4.7 & 10 \\
			\hline
			& Base & 85.39 & 99.35 & 95.35 & 84.01 & 98.89 & 94.26 & 78.90 & \textbf{99.34} & \textbf{97.44} & 72.06 & 98.92 & 96.39 & 4.0 & 6\\
			& \cellcolor{white} Ext.1 & \cellcolor[gray]{0.8} 88.58 & \cellcolor[gray]{0.8} 99.07 & \cellcolor[gray]{0.8} 97.11 & 85.03 & 98.29 & 97.85 & \cellcolor[gray]{0.8}79.80 & \cellcolor[gray]{0.8}99.12 & \cellcolor[gray]{0.8}97.12 & \cellcolor[gray]{0.8}77.02 & 	\cellcolor[gray]{0.8}98.56 & \cellcolor[gray]{0.8}97.00 & 10.2 & 3 \\
			\multirow{-3}{*}{Diff.1}& Ext.2 & 88.14 & 97.86 & 98.10 & \cellcolor[gray]{0.8}86.67 & \cellcolor[gray]{0.8}98.33 & \cellcolor[gray]{0.8}97.12 & 79.96 & 98.89 & 95.99 & 74.41 & 97.92 & 96.28 & 5.7 & 4\\
			\hline
			& Base & 85.37 & 99.39 & 95.55 & 81.60 & 99.09 & 94.45 & 73.37 & 99.30 & 97.39 & 69.91 & \textbf{99.00} & 96.57 & 4.3 & 8 \\
			\rowcolor[gray]{0.8}\cellcolor{white}&Ext.1  & \textbf{91.02} & \textbf{99.69} & \textbf{99.47} & 88.36 & \textbf{99.28} & \textbf{98.41} & \textbf{82.97} & 98.81 & 97.37 & \textbf{79.91} & 98.37 & 96.73 & \cellcolor{white}\textcolor{red}{12.1} &\cellcolor{white} 1 \\
			\multirow{-3}{*}{Diff.2}& Ext.2 & \itshape90.31 & \itshape98.53 & \itshape98.97 &\itshape\textbf{88.89} & \itshape98.91 & \itshape97.83 & \itshape81.09 & \itshape99.02 & \itshape97.01 & \itshape77.43 & \itshape98.11 & \itshape96.97 & \itshape6.1 &  \itshape2 \\
			\hline
		\end{tabular}
	\end{center}
	\vspace{-5pt}
	\caption{The OGM prediction accuracy measured by \%True-Positive (\%TP), \%True-Negative (TN) and S\textsubscript{100} (100$\times$[standard SSIM]) measures. The decreasing order of prediction accuracy is listed under ``Ord.", where 1 corresponds to the best overall accuracy. The bold-faced numbers correspond to the best performance on each column. Within each 3$\times$3 block, the shaded row corresponds to the best performance over the block. The ``t/f" column lists the time it takes to generate one frame of prediction using GTX 1080Ti GPU. }
	\label{tab:Results}
	\vspace{-10pt}
\end{table*}
\normalsize

The encoders' architecture is identical across the models; they produce $64\times64\times32$ features from $256\times256\times n$ inputs, where $n=1$ for the OGM and $n=2$ for motion-flow related features. The decoders are also similar across the models; they convert back the $64\times64\times32$ tensors to $256\times256\times n$, where $n=1$ for Diff.1 and Diff.2 models, and $n=2$ for ED and ED-Di models. Also, the decoders in Diff.1 and Diff.2 models employ a $\tanh()$ activation function on their last layer to produce compensation matrix values in $[-1,1]$ while the decoders in ED and ED-Di models employ a linear activation to produce logits. The Core RNN module consists of 4 ConvLSTM layers, where dilation at the $k$\textsuperscript{th} ConvLSTM layer is equal to $k$ for the ED-Di model and one for the rest. Therefore, in our settings, the ED-Di model at \#4 row in Table~\ref{tab:Models} is an attempt to replicate the model proposed in~\cite{Dequaire2018}. Models \#5 and \#6 represent our suggested improvements on \#4. All of the models are trained to predict one-step-ahead during the init-phase.

Table~\ref{tab:Results} lists the results. We evaluate the prediction accuracy using \%True-Positive (TP), \%True-Negative (TN) and S\textsubscript{100} measures. The S\textsubscript{100} is the standard SSIM measure multiplied by 100, in order to be on the same scale as TP and TN. The evaluation horizon is either the entire sequence (20 frames) listed under ``Whole seq.", or only over the prediction length (the second 10 frames) listed under ``Prediction seq.". Also, to evaluate the prediction accuracy over the areas occupied by the objects only, we employ the tracklets in the KITTI dataset to mask the target and the predicted OGMs and then compute the measures. The obtained results are listed under ``Objects only" and partially reflect the prediction accuracy of dynamic objects. Recall that the labels and/or tracklets are not used for training.

Over each 3-by-3 block the shaded row highlights the result with the highest sum (TP+TN+S\textsubscript{100}), reflecting the best performance among the three corresponding configurations. The last column (titled Ord.) lists the order at which the prediction accuracy decreases, that is, number 1 corresponds to the best and 12 to the worst performance, respectively.

Based on the results, our proposed difference learning model, using Farneback algorithm to extract motion-flow features, (model \#11) outperforms the other models, however, it is also computationally the most expensive one. If we switch to the two-channel difference method (model \#12), we gain a considerable improvement over the computation time, while losing a small amount of accuracy. Considering online deployment, model \#12 can be readily employed onboard of an autonomous vehicle as it takes about 60ms to predict OGM over 1 second into the future. Additionally, it is clear that inclusion of the motion-related features improves the prediction accuracy for all of the models we consider in this work. 

\begin{figure*}[]
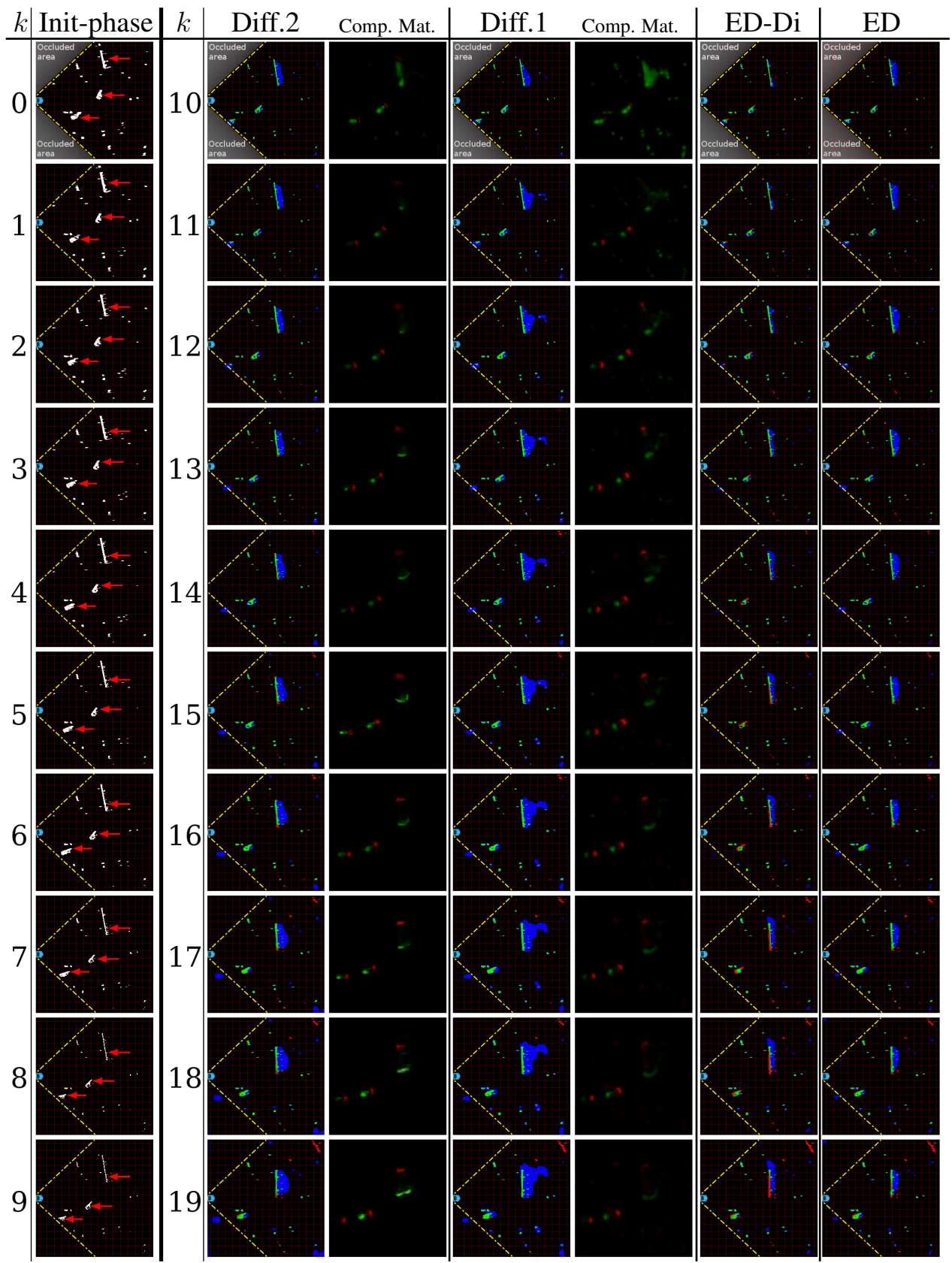

	\centering
	\ResultsAll
	\caption{Qualitative results of the four models in their best configuration. The numbers below $k$ indicate time-step. }
	\label{fig:ResultsAll}
\end{figure*}

\begin{figure*}[]
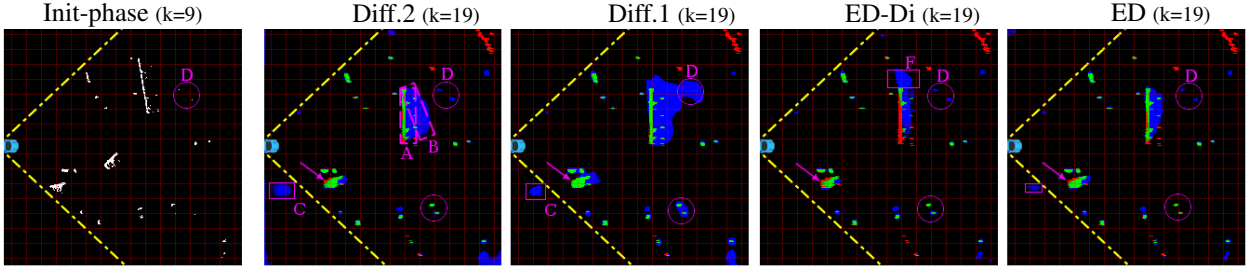

	\vspace{-5pt}
	\centering
	\ResultLastFrame
	\caption{Last frame of OGM prediction extracted from Figure~\ref{fig:ResultsAll} along with the last frame of initial OGMs.}
	\label{fig:ResultsLastFrame}
	\vspace{-10pt}
\end{figure*}

To study the predicted OGMs qualitatively, Figure~\ref{fig:ResultsAll} illustrates 10 frames of prediction generated by the four models in their best configuration. A dark red grid is overlaid on every OGM as a reference for the inertial coordinate system. Also, the dashed yellow lines show the border of the FOV. The initial OGMs (under the column ``Init-phase") are binary matrices. Therefore, they are illustrated as black and white images. The Init-phase OGMs indicate that there are three moving objects (two cars followed by a bus), highlighted by red arrows. The two cars are turning to their right. However, by the end of the Init-phase, it is difficult to determine the turning direction for the bus. 

The generated compensation matrix from the two difference learning models (Diff.1 and Diff.2) are illustrated under ``Comp. Mat."  for each of the two models. In the compensation matrices, green and red pixels correspond to numbers in $(0,1]$ and $[-1,0)$, respectively, and black pixels correspond to 0. To assess the prediction quality, the pixels of predicted OGMs are color coded as green, blue and red which correspond to true positive, false positive and false negative results. 

The qualitative results confirm the superiority of Diff.2 model over the rest; the moving objects are tracked precisely and occlusions are handled properly. To be more precise, Figure~\ref{fig:ResultsLastFrame} shows the last predicted frame extracted from Figure~\ref{fig:ResultsAll}. An important factor to notice is the inadequacy of the ground truth as a target which the prediction should be compared against. In fact, an OGM only shows the \emph{visible} border of the object. Therefore, a false positive is not undesirable everywhere on the OGMs. Over the predicted OGM by Diff.2 in Figure~\ref{fig:ResultsLastFrame}, the areas highlighted by the dashed pink rectangles (A and B) show a relatively large area of false positive. Rectangle A shows the border of the bus, grabbed from the tracklets. The ground truth OGM (green and red pixels together) only provides the visible border of the bus, however, intuitively we can say the area inside the rectangle A is fully occupied. In this case, the false positive over A is desirable. The B rectangle shows the other possible direction of the bus.

The box, labeled C, illustrates another area where a false positive is desirable. In fact, this box is out of the FOV where no visible ground truth is provided. However, the network is managed to track the moving object out of the FOV and into the occluded area. In Diff.1 predicted OGM, however, some of the false positive areas are likely undesirable. For instance, the circle D on Diff.1 OGM shows that the areas between two static objects (notice the circle D on the Init-phase OGM) and the bus are joined and the network has failed to properly predict occluded parts of the scene. 

The predicted OGM from ED and ED-Di models are considerably less accurate than Diff.2 and Diff.1. Particularly, the area under the rectangle F (on ED-Di) is wrongly predicted occupied. Also, the occupancies corresponding to the moving bus is not accurately predicted as indicated by the red areas. The ED result is slightly better than ED-Di, however, the occluded areas are not handled as well as Diff.2 model. Diff.1 model is a slightly more ``cautious" predictor, among others, due to the more false positive it generated. One interesting difference is highlighted by the arrow, where the object's border is only properly determined by Diff.1 model.

Interesting patterns emerge from the compensation matrices. For instance, looking at the compensation matrices, on Figure~\ref{fig:ResultsAll}, for the Diff.2 model over $k>11$, we may infer that there are three moving objects in the scene, two of which are smaller in length. In fact, when the compensation matrix is applied to an OGM, a green area followed by a red area is a pattern which encourages the occupancies in those areas to move from the red area towards the green area. Therefore, both the dynamic objects and their approximate headings may be interpreted from the compensation matrix. However, since the compensation matrix is a by-product of our approach, in this work we opt not to employ any inference on the compensation matrix. Illustrating the compensation matrix, however, is a supporting evidence for the difference learning idea.  

\section{Conclusion}
\label{sec:V}
Accurate multi-step prediction of OGMs, as representations of the future drivable space, is useful for path planning algorithms and does not require multi stages of engineered features in the classic approach of object detection and tracking. In this work, a difference learning architecture, based on RNNs, is proposed to predict OGMs multi-steps into the future. We have shown that our proposed architecture outperforms the state of the art in OGM multi-step prediction and is accurate in predicting the static and moving objects as well as handling occlusions. Furthermore, as a future work, the generated features by the compensation matrix from our proposed scheme may provide interesting features for label-less detection and tracking of dynamic objects.

{\small
\bibliographystyle{ieee}
\bibliography{ogm_prediction_bib}
}

\end{document}

%% file: figures.tex
\usepackage{tikz}
\usetikzlibrary{patterns,shapes.arrows,arrows,decorations.pathmorphing,backgrounds,positioning,fit,petri,matrix,decorations.pathreplacing,shapes.geometric}
\usepackage[many]{tcolorbox}

\tikzset{
Box/.style={rectangle,,thick, draw=blue!50!black!50,top color=white,bottom color=blue!50!black!20,inner sep=3pt},
Conv/.style={rectangle,,thick, draw=blue!50!black!50,top color=white,bottom color=blue!50!black!20,inner sep=1pt},
sig/.style={circle,draw=black!50,fill=blue!20,thin,inner sep=0pt,minimum size=1mm},
Sig/.style={circle,draw=black!50,fill=blue!20,thick,inner sep=0pt,minimum size=2mm},
NN/.style={rectangle,draw=black!50,fill=green!20,thick,inner sep=0pt,minimum size=6mm}, 
Func/.style={circle,draw=black!50,fill=yellow!20,thick,inner sep=0pt,minimum size=5mm}, 
Sigma/.style={circle,draw=black!50,fill=yellow!20,thick,inner sep=0pt,minimum size=1mm}, 
Delay/.style={rectangle,draw=black!50,fill=yellow!20,thick,inner sep=0pt,minimum size=7mm},
delay/.style={rectangle,draw=black!50,fill=orange!40,thin,inner sep=0pt,minimum size=3mm},
PE/.style={circle,draw=black!50,fill=red!20,thick,inner sep=0pt,minimum size=7mm},
pre/.style={<-,>=latex,thick},
post/.style={->,>=latex,thick},
conn/.style={-,>=latex,thick},
controllerblock/.style={rectangle,thick,draw=gray!50!black!50,top color=white,bottom color=orange!90!black!20},
NNblock/.style={rectangle,minimum size=7mm,thick,draw=gray!50!black!50,top color=white,bottom color=blue!50!black!20},
plantblock/.style={rectangle,minimum size=7mm,thick,draw=gray!50!black!50,top color=white,bottom color=green!50!black!20},
terminal/.style={rectangle,minimum size=6mm,rounded corners=3mm,thick,draw=black!50,top color=white,bottom color=black!20},
pics/Encoder/.style n args={2}{
	code = {
		\draw [fill==blue!50!black!50,top color=white,bottom color=blue!50!black!20,inner sep=7pt] 
		(#1,#2) -- (#1,#2+0.6) -- (#1+0.7,#2+0.2) -- (#1+0.7,#2-0.2) -- (#1,#2-0.6) -- cycle;
		\node []at (#1+0.35,#2){\small{En.}};
	}
},
pics/Decoder/.style n args={2}{
	code = {
		\draw [fill==blue!50!black!50,top color=white,bottom color=blue!50!black!20,inner sep=7pt] 
		(#1,#2) -- (#1,#2+0.2) -- (#1+0.7,#2+0.6) -- (#1+0.7,#2-0.6) -- (#1,#2-0.2) -- cycle ;
		\node []at (#1+0.35,#2){\small{De.}};
		
	}
},
pics/OGMtiny/.style n args={4}{
	code = {
		\draw [fill overzoom image*={clip,trim=0 0cm 0 0}{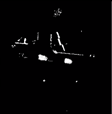},inner sep=7pt] 
		(#1,#2) -- (#1+#3,#2+#4) -- (#1+#3,#2-#4) -- (#1,#2-#4-#4) -- cycle ;
		
	}
},
pics/OGMtinyEnd/.style n args={4}{
	code = {
		\draw [fill overzoom image*={clip,trim=0 0cm 0 0}{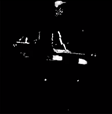},inner sep=7pt] 
		(#1,#2) -- (#1+#3,#2+#4) -- (#1+#3,#2-#4) -- (#1,#2-#4-#4) -- cycle ;
	}
}
}

\newcommand{\ClassicOBJTracking}{
	\begin{tikzpicture}
	\footnotesize
	\node [Box,text width=1.4cm, text centered] (OD) at (-0.2,-0.65)  {Object Detection};
	\node [Box,text width=1.4cm, text centered] (OC) at (2,-1.3)  {Object Classifier};
	\node [Box,text width=1.5cm,text centered] (DA) at (2,0.0)  {Data Association};
	\node [Box,text width=1.55cm,text centered] (MA) at (4.1,0.0)  {Model Assignment};
	\node [Box,text width=1.1cm,text centered] (SU) at (4.1,-1.3)  {State update};
	\node [Box,text width=1.2cm,text centered] (CSU) at (6,-0.8)  {Drivable space};
	\node at (-0.2,0.3){data};
	\draw [post, line width=0.8pt] (-0.2,0.1) -- (OD.north) ;
	\draw [post, line width=0.8pt] (OD.east) -- (OC.west);
	\draw [post, line width=0.8pt] (OD.east) -- (DA.west);
	\draw [post, line width=0.8pt] (OC) -- (DA);
	\draw [post, line width=0.8pt] (DA) -- (MA);
	\draw [post, line width=0.8pt] (MA) -- (SU);
	\draw [post, line width=0.8pt] (SU.east) -- (CSU.west);
	\draw [dashed,post,thin,line width=0.5pt] (OC.north east) -- (MA.south west);
	\draw [dashed,post,thin,line width=0.5pt] (OC.east) -- (SU);
	\draw [dashed,post,thin,line width=0.5pt] (MA.east) -| (CSU);
	\node at (6,-1.75){to path planning};
	\draw [post, line width=0.8pt](CSU.south) -- (6,-1.6);
	\end{tikzpicture}
}
\newcommand{\PredictionOBJTracking}{
	\begin{tikzpicture}
	\footnotesize
	\node [Box,text width=1.4cm, text centered] (CO) at (0,0)  {Convert to OGM};
	\node [Box,text width=1.4cm, text centered] (OMSP) at (2,0)  {OGM Multi-Step Prediction};
	\pic [local bounding box=O1]  {OGMtiny={3.5}{0.2}{0.3}{0.3}};
	\pic [local bounding box=O2]  {OGMtiny={3.9}{0.2}{0.3}{0.3}};
	\node[]at (4.6,0){\small{$\dots$}};
	\pic [local bounding box=O3]  {OGMtinyEnd={4.9}{0.2}{0.3}{0.3}};
	\draw [decorate,decoration={brace,amplitude=5pt}] (3.4,0.5) -- (5.5,0.5);
	\draw [decorate,decoration={brace,amplitude=5pt,mirror}] (3.4,-0.5) -- (5.5,-0.5);
	\node [text width=5cm, text centered] at (4.45,0.9){Predicted OGMs};
	\node []at (-1.8,0){\small{data}};
	\draw [post, line width=0.8pt] (-1.4,0) -- (CO.west) ;
	\draw [post, line width=0.8pt] (CO.east) -- (OMSP.west) ;
	\draw [post, line width=0.8pt] (OMSP.east) -- (3.3,0);
	\node at (4.5,-1.2){to path planning};
	\draw [post, line width=0.8pt](4.45, -.7) -- (4.45,-1.);
	\end{tikzpicture}
}

\newcommand{\BaseArchExtensions}{
	\begin{tikzpicture}
		\footnotesize
		\node[Sig]    (O)       at (-0.5,0)   [label=left:\small{$\mathbf{\hat{O}}_k$}]{};
		\node[Sig]    (Ostar)   at (-0.5,0.75)   [label=left:\small{$\mathbf{O}^*_k$}]{};
		\node[Sig]    (Ohat)    at (0.25,0.35)   [label=above right:\small{$\mathbf{O}_k$}]{};
		\node[Sig]    (Opred)   at (6.4,0.35)   [label=above:\small{$\mathbf{\hat{O}}_{k+1}$}]{};
		\pic [local bounding box=En1]  {Encoder={1.5}{0.35}};
		\pic [local bounding box=En2, rotate=270] {Encoder={-1.5}{2.7}};
		\pic [local bounding box=De] {Decoder={5.2}{0.35}};
		\node [Box,minimum height=0.5cm,text width=0.95cm, text centered] (MF) at (1.7,1.9)  {\footnotesize{MFE}};
		\node [Box,minimum height=0.5cm, text width=0.6cm, text centered] (RNN) at (4,0.35)  {\footnotesize{Core RNN}};
		\node [Sigma,text width=0.3cm,text centered] (STACK)  at (2.7,0.35) []{\footnotesize{S}};
		\node [delay,text width=0.4cm,text centered] (Delay1) at (3.1,-0.5)  {D};
		\node [delay,text width=0.4cm,text centered] (Delay2) at (0.25,1.9)  {D};
		\node at (0,0.9){$\boldsymbol{\delta}_k$};
		\draw [conn, line width=1pt] (Ostar) -- (Ohat);
		\draw [dashed,>=latex,thin,line width=0.5pt] (O) -- (Ohat);
		\draw [post, line width=1pt] (Ohat) -- (Delay2);
		\draw [post, line width=1pt] (Delay2) -- (MF);
		\draw [post, line width=1pt] (0.25, 1.2) -- (1, 1.2) -- (MF);
		\draw [post, line width=1pt] (MF) -- (2.7, 1.9) -- (En2);
		\draw [post, line width=1pt] (En2)-- (STACK); 
		\draw [post, line width=1pt] (Ohat) -- (En1);
		\draw [post, line width=1pt] (En1) -- (STACK);
		\draw [post, line width=1pt] (STACK) -- (RNN);
		\draw [post, line width=1pt] (RNN) -- (De);
		\draw [post, line width=1pt] (De) -- (Opred);
		\draw [post, line width=1pt] (Opred) -- (6.4,-0.5)--(Delay1);
		\draw [post, line width=1pt] (Delay1) -- (-0.5,-0.5)--(O);
		\draw [post, line width=1pt] (0.0, 0.7) arc (145:210:0.8) ;
		\node [delay,text width=0.4cm,text centered] (DelayL) at (5,2)  {D};
		\node at(5.75,1.98) {: Delay};
		\node [Sigma,text width=0.3cm,text centered] (STACKL)  at (5,1.5) []{\footnotesize{S}};
		\node at(5.73,1.48) {: Stack};
		
	\end{tikzpicture}
}

\newcommand{\DiffLearningBlockDiagram}{
	\begin{tikzpicture}
	\footnotesize
	\node [Box,text width=1.cm,text centered] (DL) at (1.7,0.35)  {Diff. Learn};
	\node [Box,text width=1.2cm,text centered] (OC) at (5,0.35)  {Classifier};
	\node [Sigma,text width=0.3cm,text centered] (Sum) at (3.5,0.35)  {$+$};
	\node [delay,text width=0.4cm,text centered] (Delay) at (3.1,-0.5)  {D};
	
	\node[Sig]    (O)       at (-0.5,0)   [label=left:\small{$\mathbf{\hat{O}}_k$}]{};
	\node[Sig]    (Ostar)   at (-0.5,0.75)   [label=left:\small{$\mathbf{O}^*_k$}]{};
	\node[Sig]    (Ohat)    at (0.25,0.35)   [label=above right:\small{$\mathbf{O}_k$}]{};
	\node[Sig]    (Opred)    at (6.4,0.35)   [label=above:\small{$\mathbf{\hat{O}}_{k+1}$}]{};
	\node at (0,0.9){\small{$\boldsymbol{\delta}_k$}};
	\node at (2.9,0.6){\small{$\Delta\mathbf{O}_k$}};
	
	\draw [conn, line width=1pt] (Ohat) -- (Ostar);
	\draw [dashed,>=latex,thin,line width=0.5pt] (O) -- (Ohat);
	\draw [post, line width=1pt] (Ohat) -- (DL);
	\draw [post, line width=1pt] (Ohat) -- (0.25, 1) -- (3.5, 1) -- (Sum);
	\draw [post, line width=1pt] (DL) -- (Sum);
	\draw [post, line width=1pt] (Sum) -- (OC);
	\draw [post, line width=1pt] (Opred) -- (6.4,-0.5)--(Delay);
	\draw [post, line width=1pt] (OC) -- (Opred);
	\draw [post, line width=1pt] (Delay) -- (-0.5,-0.5)--(O);
	\draw [post, line width=1pt]  (0.0, 0.7) arc (145:210:0.8) ;
	\end{tikzpicture}
}

\newcommand{\DiffLearningBlockDiagramFull}{
	\begin{tikzpicture}
		\footnotesize
		\fill [top color=red!5, bottom color=red!10] 
		(0,2.3) -- (4.7, 2.3) --(4.7,-0.4) -- (0, -0.4)-- cycle;
		\node[Sig]    (O)       at (-0.3,0)   [label=left:\small{$\mathbf{\hat{O}}_k$}]{};
		\node[Sig]    (Ostar)   at (-0.3,0.75)   [label=left:\small{$\mathbf{O}^*_k$}]{};
		\node[Sig]    (Ohat)    at (0.25,0.35)   [label=above right:\small{$\mathbf{O}_k$}]{};
		\node[Sig]    (OComp)   at (5,0.35)   [label=right:\small{$\Delta\mathbf{O}_{k+1}$}]{};
		\node[Sig]    (OPred)   at (7,-0.5)   [label=above:\small{$\mathbf{\hat{O}}_{k+1}$}]{};
		\pic [local bounding box=En1]  {Encoder={1.}{0.35}};
		\pic [local bounding box=En2, rotate=270] {Encoder={-1.55}{2.2}};
		\pic [local bounding box=De] {Decoder={3.85}{0.35}};
		\node [Box,minimum height=0.5cm,text width=0.7cm, text centered] (MF) at (1.4,1.9)  {\footnotesize{MFE}};
		\node [Box,minimum height=0.5cm, text width=0.6cm, text centered] (RNN) at (3.1,0.35)  {\footnotesize{Core RNN}};
		\node [Sigma,text width=0.3cm,text centered] (STACK)  at (2.2,0.35) []{\footnotesize{S}};
		\node [Sigma,text width=0.3cm,text centered] (SUM)  at (5,-0.5) []{$+$};
		\node [Box,text width=0.8cm,text centered] (OC) at (6,-0.5)  {Classi-fier};
		\node [delay,text width=0.4cm,text centered] (Delay1) at (3.1,-1.)  {D};
		\node [delay,text width=0.4cm,text centered] (Delay2) at (0.25,1.9)  {D};
		\node at (0,0.9){$\boldsymbol{\delta}_k$};
		\draw [conn, line width=1pt] (Ostar) -- (Ohat);
		\draw [dashed,>=latex,thin,line width=0.5pt] (O) -- (Ohat);
		\draw [post, line width=1pt] (Ohat) -- (Delay2);
		\draw [post, line width=1pt] (Delay2) -- (MF);
		\draw [post, line width=1pt] (0.25, 1.2) -- (1, 1.2) -- (MF);
		\draw [post, line width=1pt] (MF) -| (En2);
		\draw [post, line width=1pt] (En2)-- (STACK); 
		\draw [post, line width=1pt] (Ohat) -- (En1);
		\draw [post, line width=1pt] (En1) -- (STACK);
		\draw [post, line width=1pt] (STACK) -- (RNN);
		\draw [post, line width=1pt] (RNN) -- (De);
		\draw [post, line width=1pt] (De) -- (OComp);
		\draw [post, line width=1pt] (Ohat) |- (SUM);
		\draw [post, line width=1pt] (OComp) -- (SUM);
		\draw [post, line width=1pt] (SUM) -- (OC);
		\draw [post, line width=1pt] (OC) -- (OPred);
		\draw [post, line width=1pt] (OPred) |- (Delay1);
		\draw [post, line width=1pt] (Delay1) -| (O);
		\draw [post, line width=1pt] (0.1, 0.7) arc (145:210:0.8) ;	
	\end{tikzpicture}
}

\newcommand{\ResultsAll}{
	\begin{tikzpicture}
	    \node[anchor=south west,inner sep=0] at (0,0) {\includegraphics[scale=0.23,trim=0mm 0mm 0mm 24mm,clip]{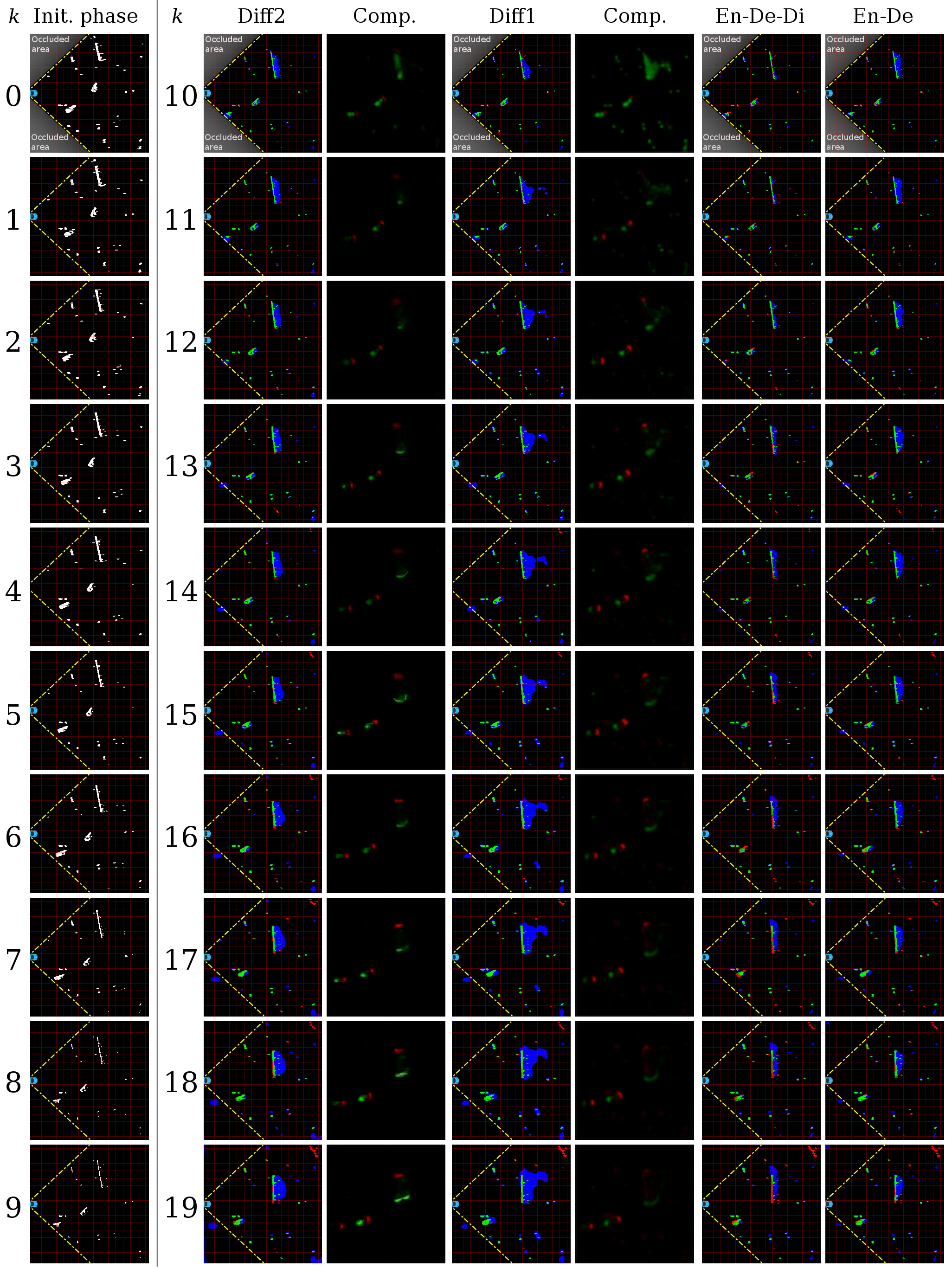}};
		\node[] at (0.27,21.95)  {\Large$k$};
		\node[] at (1.6,21.9)  {\Large Init-phase};
		\node[] at (3.15,21.95)  {\Large$k$};
		\node[] at (4.6,21.95)  {\Large Diff.2};
		\node[] at (6.75,21.85)  {Comp. Mat.};
		\node[] at (9,21.95)  {\Large Diff.1};
		\node[] at (11.05,21.85)  {Comp. Mat.};
		\node[] at (13.4,21.95)  {\Large ED-Di};
		\node[] at (15.5,21.95)  {\Large ED};
		\draw [conn, line width=1pt] (0,21.65) -- (16.7,21.65);
		\draw [line width=0.5pt] (0.45,0) -- (0.45,22.2);
		\draw [line width=2pt] (2.75,0) -- (2.75,22.2);
		\draw [line width=0.5pt] (3.5,0) -- (3.5,22.2);
		\draw [line width=1pt] (7.85,0) -- (7.85,22.2);
		\draw [line width=1pt] (12.222,0) -- (12.222,22.2);
		\draw [line width=1pt] (14.405,0) -- (14.405,22.2);
		
		\foreach \y in {0,...,9}
			\draw [post, color=red] (2.2,21.3-\y*2.2) -- (1.8,21.3-\y*2.2);
		\foreach \y in {0,...,7}
			\draw [post, color=red] (2.1-\y*\y*0.002,20.65-\y*2.15 - \y*\y*0.005) -- (1.7-\y*\y*0.002,20.65-\y*2.15- \y*\y*0.005);
		\foreach \y in {8,9}
			\draw [post, color=red] (2.05-\y*\y*0.002,20.65-\y*2.15 - \y*\y*0.004) -- (1.65-\y*\y*0.002,20.65-\y*2.15- \y*\y*0.004);
		\foreach \y in {0,...,5}
			\draw [post, color=red] (1.65-\y*\y*0.0035,20.25-\y*2.15 - \y*\y*0.003) -- (1.3-\y*\y*0.0035,20.25-\y*2.15- \y*\y*0.003);
		\foreach \y in {6,...,9}
			\draw [post, color=red] (1.65-\y*\y*0.0035,20.3-\y*2.15 - \y*\y*0.0025) -- (1.3-\y*\y*0.0035,20.3-\y*2.15- \y*\y*0.0025);
		
	\end{tikzpicture}
}
\newcommand{\ResultLastFrame}{
	\begin{tikzpicture}
		\node[anchor=south west,inner sep=0] at (0,0) {\includegraphics[scale=0.35,trim=0mm 0mm 0mm 13mm,clip]{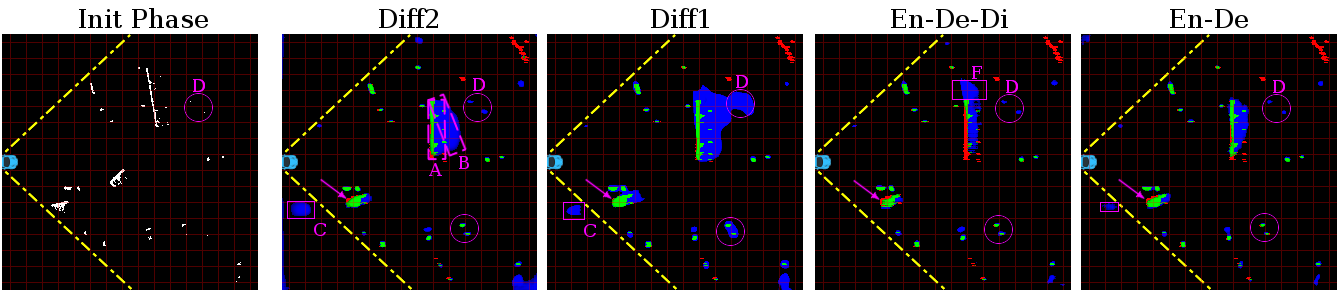}};
		\node[] at (1.6,3.4)  { Init-phase \footnotesize{(k=9)}};
		\node[] at (5.5,3.4)  { Diff.2 \footnotesize{(k=19)}};
		\node[] at (8.9,3.4)  { Diff.1 \footnotesize{(k=19)}};
		\node[] at (12.1,3.4)  { ED-Di \footnotesize{(k=19)}};
		\node[] at (15.4,3.4)  { ED \footnotesize{(k=19)}};
	\end{tikzpicture}
}

%% file: ogm_prediction.bbl
\begin{thebibliography}{10}\itemsep=-1pt

\bibitem{Cho2014}
K.~Cho, B.~Van~Merri{\"e}nboer, C.~Gulcehre, D.~Bahdanau, F.~Bougares,
  H.~Schwenk, and Y.~Bengio.
\newblock Learning phrase representations using rnn encoder-decoder for
  statistical machine translation.
\newblock {\em arXiv preprint arXiv:1406.1078}, 2014.

\bibitem{Dequaire2018}
J.~Dequaire, P.~Ondrúška, D.~Rao, D.~Wang, and I.~Posner.
\newblock Deep tracking in the wild: End-to-end tracking using recurrent neural
  networks.
\newblock {\em The International Journal of Robotics Research},
  37(4-5):492--512, 2018.

\bibitem{Elfes1990}
A.~Elfes.
\newblock Occupancy grids: A stochastic spatial representation for active robot
  perception.
\newblock In {\em Proceedings of the Sixth Conference on Uncertainty in AI},
  volume 2929, page~6, 1990.

\bibitem{Engel2018}
N.~Engel, S.~Hoermann, P.~Henzler, and K.~Dietmayer.
\newblock Deep object tracking on dynamic occupancy grid maps using rnns.
\newblock {\em arXiv preprint arXiv:1805.08986}, 2018.

\bibitem{Ess2010}
A.~Ess, K.~Schindler, B.~Leibe, and L.~Van~Gool.
\newblock Object detection and tracking for autonomous navigation in dynamic
  environments.
\newblock {\em The International Journal of Robotics Research},
  29(14):1707--1725, 2010.

\bibitem{Farneback2003}
G.~Farneb{\"a}ck.
\newblock Two-frame motion estimation based on polynomial expansion.
\newblock In {\em Scandinavian conference on Image analysis}, pages 363--370.
  Springer, 2003.

\bibitem{Finn2016}
C.~Finn, I.~Goodfellow, and S.~Levine.
\newblock Unsupervised learning for physical interaction through video
  prediction.
\newblock In {\em Advances in neural information processing systems}, pages
  64--72, 2016.

\bibitem{Geiger2013}
A.~Geiger, P.~Lenz, C.~Stiller, and R.~Urtasun.
\newblock Vision meets robotics: The kitti dataset.
\newblock {\em International Journal of Robotics Research (IJRR)}, 2013.

\bibitem{Gupta2008}
O.~K. Gupta and R.~A. Jarvis.
\newblock Optimal global path planning in time varying environments based on a
  cost evaluation function.
\newblock In {\em Australasian Joint Conference on Artificial Intelligence},
  pages 150--156. Springer, 2008.

\bibitem{Hoermann2017}
S.~Hoermann, M.~Bach, and K.~Dietmayer.
\newblock Dynamic occupancy grid prediction for urban autonomous driving: A
  deep learning approach with fully automatic labeling.
\newblock {\em arXiv preprint arXiv:1705.08781}, 2017.

\bibitem{Jaderberg2015}
M.~Jaderberg, K.~Simonyan, A.~Zisserman, et~al.
\newblock Spatial transformer networks.
\newblock In {\em Advances in neural information processing systems}, pages
  2017--2025, 2015.

\bibitem{Lotter2016}
W.~Lotter, G.~Kreiman, and D.~Cox.
\newblock Deep predictive coding networks for video prediction and unsupervised
  learning.
\newblock {\em arXiv preprint arXiv:1605.08104}, 2016.

\bibitem{Mathieu2015}
M.~Mathieu, C.~Couprie, and Y.~LeCun.
\newblock Deep multi-scale video prediction beyond mean square error.
\newblock {\em arXiv preprint arXiv:1511.05440}, 2015.

\bibitem{Miller2007}
I.~Miller and M.~Campbell.
\newblock A mixture-model based algorithm for real-time terrain estimation.
\newblock In {\em The 2005 DARPA Grand Challenge}, pages 407--436. Springer,
  2007.

\bibitem{Mohajerin2017_2}
N.~Mohajerin, M.~Mozifian, and S.~L. Waslander.
\newblock Deep learning a quadrotor dynamic model for multi-step prediction.
\newblock In {\em Robotics and Automation (ICRA), IEEE International Conference
  on}. IEEE, 2018.

\bibitem{Mohajerin2017_1}
N.~Mohajerin and S.~L. Waslander.
\newblock State initialization for recurrent neural network modeling of
  time-series data.
\newblock In {\em Neural Networks (IJCNN), International Joint Conference on},
  pages 2330--2337. IEEE, 2017.

\bibitem{Noguchi2012}
H.~Noguchi, T.~Yamada, T.~Mori, and T.~Sato.
\newblock Mobile robot path planning using human prediction model based on
  massive trajectories.
\newblock In {\em Networked Sensing Systems (INSS), 2012 Ninth International
  Conference on}, pages 1--7. IEEE, 2012.

\bibitem{Nuss2018}
D.~Nuss, S.~Reuter, M.~Thom, T.~Yuan, G.~Krehl, M.~Maile, A.~Gern, and
  K.~Dietmayer.
\newblock A random finite set approach for dynamic occupancy grid maps with
  real-time application.
\newblock {\em The International Journal of Robotics Research}, 37(8):841--866,
  2018.

\bibitem{Ohki2010}
T.~Ohki, K.~Nagatani, and K.~Yoshida.
\newblock Collision avoidance method for mobile robot considering motion and
  personal spaces of evacuees.
\newblock In {\em Intelligent Robots and Systems (IROS), 2010 IEEE/RSJ
  International Conference on}, pages 1819--1824. IEEE, 2010.

\bibitem{Ondruska2016}
P.~Ondruska and I.~Posner.
\newblock Deep tracking: Seeing beyond seeing using recurrent neural networks.
\newblock {\em arXiv preprint arXiv:1602.00991}, 2016.

\bibitem{Petrovskaya2009}
A.~Petrovskaya and S.~Thrun.
\newblock Model based vehicle detection and tracking for autonomous urban
  driving.
\newblock {\em Autonomous Robots}, 26(2-3):123--139, 2009.

\bibitem{Santana2011}
A.~M. Santana, K.~R. Aires, R.~M. Veras, and A.~A. Medeiros.
\newblock An approach for 2d visual occupancy grid map using monocular vision.
\newblock {\em Electronic Notes in Theoretical Computer Science}, 281:175--191,
  2011.

\bibitem{Srivastava2015}
N.~Srivastava, E.~Mansimov, and R.~Salakhudinov.
\newblock Unsupervised learning of video representations using lstms.
\newblock In {\em International conference on machine learning}, pages
  843--852, 2015.

\bibitem{Tsardoulias2016}
E.~Tsardoulias, A.~Iliakopoulou, A.~Kargakos, and L.~Petrou.
\newblock A review of global path planning methods for occupancy grid maps
  regardless of obstacle density.
\newblock {\em Journal of Intelligent \& Robotic Systems}, 84(1-4):829--858,
  2016.

\bibitem{Wang2018}
S.~Wang, R.~Clark, H.~Wen, and N.~Trigoni.
\newblock End-to-end, sequence-to-sequence probabilistic visual odometry
  through deep neural networks.
\newblock {\em The International Journal of Robotics Research},
  37(4-5):513--542, 2018.

\bibitem{Wang2003}
Z.~Wang, E.~P. Simoncelli, and A.~C. Bovik.
\newblock Multiscale structural similarity for image quality assessment.
\newblock In {\em The Thrity-Seventh Asilomar Conference on Signals, Systems \&
  Computers, 2003}, volume~2, pages 1398--1402. Ieee, 2003.

\bibitem{Xingjian2015}
S.~Xingjian, Z.~Chen, H.~Wang, D.-Y. Yeung, W.-K. Wong, and W.-c. Woo.
\newblock Convolutional lstm network: A machine learning approach for
  precipitation nowcasting.
\newblock In {\em Advances in neural information processing systems}, pages
  802--810, 2015.

\end{thebibliography}
